\documentclass[accepted]{uai2023} 

\usepackage[american]{babel}

\usepackage{natbib} 
\usepackage{mathtools} 
\usepackage{booktabs} 
\usepackage{tikz} 
\usepackage{amsmath}
\usepackage{amssymb}
\usepackage{subcaption}

\newcommand{\Org}{\textsf{Org}}
\newcommand{\Mag}{\textsf{MAgent}}

\title{Latent Interactive A2C for Improved RL in Open Many-Agent Systems}

\author[1]{Keyang He}
\author[1]{Prashant Doshi}
\author[2]{Bikramjit Banerjee}

\affil[1]{%
    School of Computing\\
    University of Georgia\\
    Athens, Georgia, USA
}
\affil[2]{%
    School of Computing Sciences and Computer Engineering\\
    University of Southern Mississippi\\
    Hattiesburg, Mississippi, USA\\
}
  
\begin{document}
\maketitle

\begin{abstract}
There is a prevalence of multiagent reinforcement learning (MARL) methods that engage in centralized training. But, these methods involve obtaining various types of information from the other agents, which may not be feasible in competitive or adversarial settings. A recent method, the interactive advantage actor critic (IA2C), engages in decentralized training coupled with decentralized execution, aiming to predict the other agents' actions from possibly noisy observations. In this paper, we present the latent IA2C that utilizes an encoder-decoder architecture to learn a latent representation of the hidden state and other agents' actions. Our experiments in two domains -- each populated by many agents -- reveal that the latent IA2C significantly improves sample efficiency by reducing variance and converging faster. Additionally, we introduce open versions of these domains where the agent population may change over time, and evaluate on these instances as well.  
\end{abstract}

\section{Introduction}
\label{sec:introduction}
In the area of machine learning, reinforcement learning (RL) is the problem of an agent learning on its own what to do in various situations based on the rewards it receives as it actively explores its environment. It must contend with exploring versus exploiting and actions that often exhibit longer-term impact (delayed reward). Methods for RL~\citep{Kaelbling96:Reinforcement} are appealing because the agent self-learns control potentially without accessing any prior specification of the environment.

Recent multiagent RL (MARL) techniques have either generalized policy optimization or the actor-critic (AC) architecture to multiagent settings while predominantly utilizing centralized training and decentralized execution (CTDE). These techniques maintain a joint-action value or advantage function whereas the actor learns individual policies that map each agent's own observations to its actions. For example, multiagent proximal policy optimization (MA-PPO)~\citep{mappo} updates a centralized value function with action-observation inputs from all agents and utilizes this value function to learn improved policies for each agent. AC based methods such as MADDPG~\citep{maddpg}, LOLA~\citep{Foerster18:LOLA}, and COMA~\cite{coma} all subscribe to CTDE with the agents exchanging varying types of information to facilitate a centralized critic. However, such information exchanges may not be feasible in competitive and adversarial settings. This motivates a {\em decentralized training and decentralized execution} (DTDE) approach, as adopted by AC based methods such as IA2C~\citep{ia2c}, which takes a subjective perspective to the learning and interactions by modeling other agents. Such decentralized approaches also better align with the goal of striving for optimality, i.e., the AI stance of designing an optimal learning agent effective in its environment~\citep{Shoham07:If}.

IA2C was recently scaled to many-agent systems using Dirichlet-multinomial modeling~\citep{mia2c}. This method, labeled IA2C$^{++}$, updates the Dirichlet-multinomial model using private observations of the agent but separately from the AC. While IA2C$^{++}$ is shown to converge to high quality policies in mixed-motive settings, our experiments reveal that the method suffers from high variance, particularly in domains where certain states admit a single action as the optimal action, which the Dirichlet-multinomial model is unable to predict and also because the model's predictions often change drastically when the sample size is initially low. 

In order to improve the sample efficiency, we present a variant of IA2C$^{++}$ that integrates the Dirichlet-multinomial model and its update into our neural pipeline using an encoder-decoder architecture, which also leads to an end-to-end neural network architecture for this DTDE technique. Latent in the encoder-decoder is a representation of the hidden state and the other agents' mean-field action, which is input to the advantage value function in the AC. Use of this latent variable in the critic additionally has a regularizing effect, leading to lower variance and improved sample complexity, which has been corroborated in recent work~\citep{slac,side}.   


We evaluate this new method in two domains, \Org{}~\citep{ia2c} and \Mag{}~\citep{magent}, with up to a hundred agents in the domains. \Org{} models a typical business organization featuring a mix of cooperation for achieving the overall improvement of the organization and individual employee competition. In this paper, we let \Org{} be an {\em open} many-agent system where the supervisory agents may add or remove employees; agent openness is thus controlled by the agents within the system instead of being driven extraneously. \Mag{} is a battlefield simulation game featuring two groups of agents for which we also introduce an open variant. For both these domains, we find that the latent IA2C exhibits less variance and is significantly more sample efficient than relevant baselines.

\section{Background: RL under Partial Observability}
\label{sec:background}

Non-stationary environments are a primary challenge for MARL. In cooperative settings, agents may directly share their policies with others. In competitive settings, this may not be possible and inferring other agents' actions based on their past action histories is a common workaround. In real-world problems, where communication between agents are not possible and other agents' actions are not perfectly observable, frequentist inference methods such as maximum likelihood estimation struggles to provide accurate predictions. 

A recent RL method, Interactive A2C (IA2C), engages in DTDE in partially observable multiagent settings. This perspective to RL training makes it better suited for mixed cooperative-competitive environments compared to the previous MARL methods~\citep{coma,maddpg, Foerster18:LOLA} that mostly engage in CTDE, which require varied types of information exchanges between possibly competitive agents, but these may not be feasible. We briefly review IA2C below and also discuss how it has been extended to perform RL in settings shared with several agents.

\subsection{Interactive A2C}

Interactive advantage actor-critic (IA2C)~\citep{ia2c} is a decentralized AC method designed for egocentric RL in partially observable Markovian settings shared with other agents. In IA2C, each agent has its own critic and actor neural network, the former mapping individual observations to joint action values in terms of the agent's own reward function, $Q_0(o,a_0,\mathbf{a}_{-0})$, and the latter mapping individual observations to individual action probabilities, $\pi_{0,\boldsymbol{\theta}}(a_0|o)$, $\boldsymbol{\theta}$ is its set of parameters. IA2C estimates advantages as
\begin{equation*}
    A_0(o,a_0,\hat{\mathbf{a}}_{-0}) = avg [r + \gamma Q_0(o,a_0',\hat{\mathbf{a}}_{-0}') - Q_0(o,a_0,\hat{\mathbf{a}}_{-0})]
\end{equation*}
while the actor's gradient is estimated as 
\[avg[\nabla_{\boldsymbol{\theta}}\log\pi_{0,\boldsymbol{\theta}}(a_0|o)~A_0(o,a_0,\hat{\mathbf{a}}_{-0})]\]
where $r,o$ and $a_0'$ are samples from the trajectory, $\hat{\mathbf{a}}_{-0}$ and $\hat{\mathbf{a}}_{-0}'$ are {\em predicted} actions of the other agents, and the $avg$ is taken over sampled trajectories. In contrast with previous multiagent deep RL algorithms, IA2C does not require access to other agents' actions and/or gradients. Rather, agents maintain belief distributions over other agents' possible models. Let $\mathbf{a}_{-0}=a_1\ldots,a_N$. Given the agent's prior belief $b_0$, action $a_0$, as well as its public and private observations $o_0'$, $\omega_0'$, the agent updates its belief over agent $j$'s model, $m_j'=\langle \pi_j',h_j'\rangle$, as
\begin{align}
    & b_0'(m_j'|b_0,a_0,o_0',\omega_0')\propto \sum_{\mathbf{a}_{-0}} 
    \biggl( \prod_{k=1,k\neq j}^N\sum_{m_k \in M_k} b_0(m_k) \nonumber\\
    &Pr(a_k|m_k)\sum_{m_j\in M_j} ~b_0(m_j)~Pr(a_j|m_j) ~\delta_K(\pi_j',\pi_j) \nonumber \\
    &\delta_K(APPEND(h_j,\langle a_j,o' \rangle), h_j')\biggr) ~W_0(a_0,\mathbf{a}_{-0},\omega_0')
\label{eq:model-bu}
\end{align}
where public and private observations are noisy observations of states and other agents' actions. $m_j$ denotes agent $j$'s model, $\pi_j$ is $j$'s policy and $h_j$ is its action-observation history. $W_0$ is the private observation function that maps joint actions to the subject agent's private observations. $\delta_K$ is the Kronecker delta function and APPEND returns a string with the second argument appended to its first. The belief update is performed by a separate belief filter integrated into the critic network.



\subsection{Many-Agent Interactive A2C}
\label{sec:IA2C$^{++}$}

Another key challenge for MARL is the exponential growth of the joint action space. Plenty of many-agent environments naturally exhibit the {\em action anonymity} feature, which implies that environment dynamics and rewards depend on {\em action configuration} (notated henceforth as ${\cal C}$), which is the count distribution of actions in the population. Many-agent IA2C (IA2C$^{++}$)~\citep{mia2c} extends IA2C by utilizing this key insight of action anonymity thereby using configurations which scale polynomially with the number of agents rather than using joint action vectors that scale exponentially. 

\begin{figure}[ht!]
    \includegraphics[width=.45\textwidth]{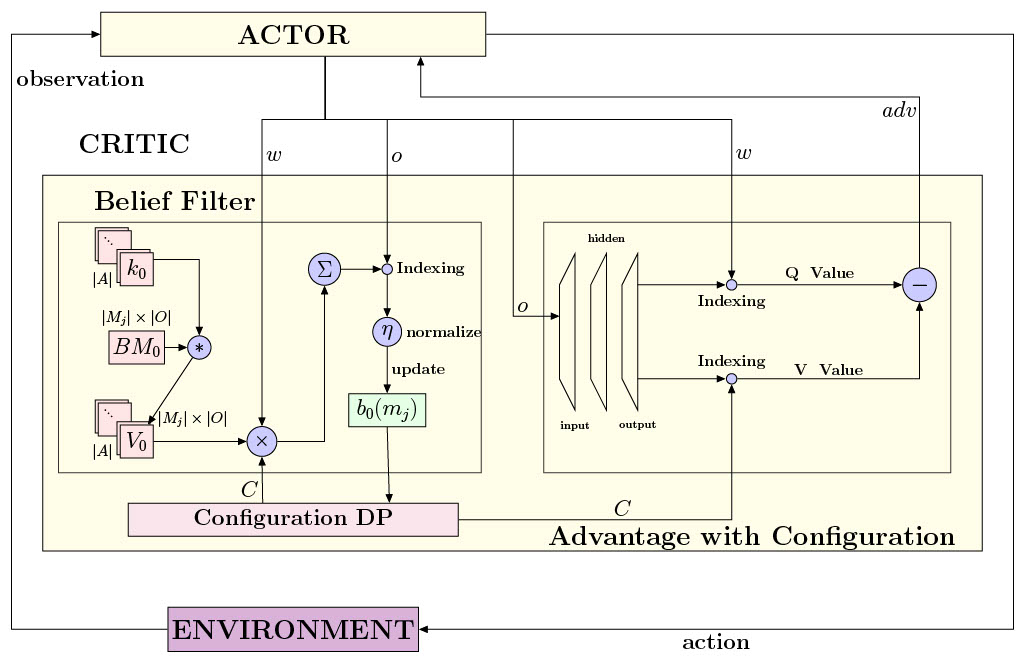}
    \caption{\cite{mia2c}'s architecture of the IA2C$^{++}$BU network. The belief filter updates the agent’s belief over the population according to observed configurations using a dynamic programming procedure.}
    \label{fig:ia2c}
\end{figure}

IA2C$^{++}$BU adapts Eq.~\ref{eq:model-bu} to update agent $0$’s belief over one other agent’s possible models in the context of action configurations instead of joint actions. 
\begin{align}
&b_0'(m_j'|b_0,a_0,o_0',\omega_0')\propto \sum\limits_{m_j \in M_j} b_0(m_j) ~ \sum\limits_{a_j} Pr(a_j|m_j)\nonumber\\
&\sum\limits_{\mathcal{C}\in\boldsymbol{\mathcal{C}}^\mathbf{a_{-0}}} Pr(\mathcal{C}|b_0(M_1),b_0(M_2),\ldots,b_0(M_N))~W_0(a_0,\mathcal{C},\omega_0')\nonumber\\
&\delta_K(\pi_j,\pi_j')\delta_K(APPEND(h_j, \langle a_j,o' \rangle),h_j').
\label{eq:model-bu-cfg}
\end{align}
The term $Pr(\mathcal{C}|b_0(M_1), b_0(M_2), \ldots, b_0(M_N))$ is the probability of configuration $\mathcal{C}$ in the distribution over the set of configurations $\boldsymbol{\mathcal{C}}^\mathbf{a_{-0}}$. The distribution is obtained using the dynamic programming procedure introduced in~\citep{configuration}. The algorithm takes as input just $N$ beliefs each of size $|M_j|$ compared to a single large belief of exponential size $|M_j|^N$. Figure~\ref{fig:ia2c} demonstrates the network architecture of IA2C$^{++}$BU.

IA2C$^{++}$BU requires a pre-defined model set for belief update. IA2C$^{++}$DM removes that assumption by using a Dirichlet-multinomial model to approximate action distributions of the entire agent population, which can then yield the mean action. Suppose the action space for the homogeneous agent population is $\{a_1, a_2, \dots, a_{|A|}\}$, and for each agent $i$, the probability of picking action $a_n$ is $\theta_n$. Let the distribution $\boldsymbol{\theta} = (\theta_1, \theta_2, \dots, \theta_{|A|})$. $\boldsymbol{\theta}$ has a Dirichlet-multinomial prior distribution with parameter $\boldsymbol{\alpha}$ if
\begin{equation}
    Pr(\boldsymbol{\theta}|\boldsymbol{\alpha}) = \frac{\Gamma(\sum_n \alpha_n)}{\Pi_n\Gamma(\alpha_n)}\Pi_n\theta_n^{\alpha_n-1}
    \label{eq:theta_given_alpha}
\end{equation}
where $\alpha_n > 0$ for all $n$, $\boldsymbol{\alpha} = (\alpha_1, \alpha_2, \dots, \alpha_N)$, and $\sum_n\theta_n = 1$. We can write this as $\boldsymbol{\theta} \sim Dir(\boldsymbol{\alpha};\boldsymbol{\theta})$. Then, probability of an action configuration $\mathcal{C}$ is expressed as:
\begin{equation}
    Pr(\mathcal{C}|\boldsymbol{\theta})=Pr(\#a_1, \#a_2, \dots, \#a_{|A|}|\boldsymbol{\theta}) = \Pi_{n=1}^{|A|} \theta_n^{\#a_n}
    \label{eq:c_given_theta}
\end{equation}
where $\#a_k$ is the number of agents selecting the $k$th action. After executing action $a_0$ and receiving (noisy) private observation $\omega_0' = (\#a_1, \#a_2, \dots, \#a_{|A|})$, the subject agent's Dirichlet-multinomial distribution can be updated by noting that the posterior $Pr(\boldsymbol{\theta}|a_0,\omega_0') $ is proportional to:
\begin{align}
    &\propto\sum_{\mathcal{C}}W_0(a_0,\mathcal{C}, \omega_0')Pr(\mathcal{C}| \boldsymbol{\theta})\cdot Dir(\boldsymbol{\alpha};\boldsymbol{\theta}) \label{eqn:step}\\
    & = \sum_{\mathcal{C}}W_0(a_0,\mathcal{C}, \omega_0')Dir(\boldsymbol{\alpha}+\mathcal{C};\boldsymbol{\theta})\label{eqn:true_posterior}\\
    &\approx Dir(\boldsymbol{\alpha}+\mathcal{C}';\boldsymbol{\theta})\label{eqn:approx_posterior}
\end{align}
Eq. \ref{eqn:step} makes a simplifying modeling assumption that the true action counts of other agents, $\mathcal{C}$, is conditionally independent of $a_0$ given $\boldsymbol{\theta}$. For the sake of tractability, the last step is approximated as a single component, $Dir(\boldsymbol{\alpha}+\mathcal{C}';\boldsymbol{\theta})$. ${\mathcal C'}$ can be calculated in several alternative ways as discussed by~\cite{mia2c}, and in this paper we use the {\em rectified} method. 

In IA2C$^{++}$, the critic network estimates advantage as:
\begin{equation*}
\resizebox{\linewidth}{!}{$
    \displaystyle
    A_0(o, a_0, \mathcal{C}^\mathbf{a_{-0}}) = avg[r + \gamma ~Q_0(o'', a_0', \mathcal{C}^\mathbf{a_{-0}'}) - Q_0(o', a_0, \mathcal{C}^\mathbf{a_{-0}})]
$}
\end{equation*}
while the actor network's gradient is:
\begin{align}
    avg \left [ \nabla_{\boldsymbol{\theta}} ~log ~\pi_{0,\boldsymbol{\theta}}(a_0|\boldsymbol{o}) ~A_0(o', a_0,\mathcal{C}^\mathbf{a_{-0}}) \right ]
\label{eq:gradient}
\end{align}
where $r$, $o'$, $o''$, and $a_0'$ are samples, $\mathbf{a_{-0}}$ and $\mathbf{a_{-0}'}$ are the predicted joint actions of the other agents for the current and next step, respectively, replaced by their corresponding configurations, and $avg$ is taken over the sampled trajectories. The predicted action for the next time step is sampled from the updated Dirichlet distribution. 

\section{Open Organization Domain}
\label{sec:org}

A typical business organization (\Org{}) features a mix of cooperation among the employees for improving the overall financial health of the organization and individual employee competition. \Org{}'s states correspond to the organization's financial health. Agents (employees) can observe the amount of orders received by the organization only, which are noisy indicators of \Org{}'s underlying states. An agent's rewards come from an individual reward component that depends on the agent's own action and a group reward component that depends on the joint action of all agents, which can lead to coordination with others. 

\begin{figure}[ht!]
    \centerline{\includegraphics[width=.45\textwidth]{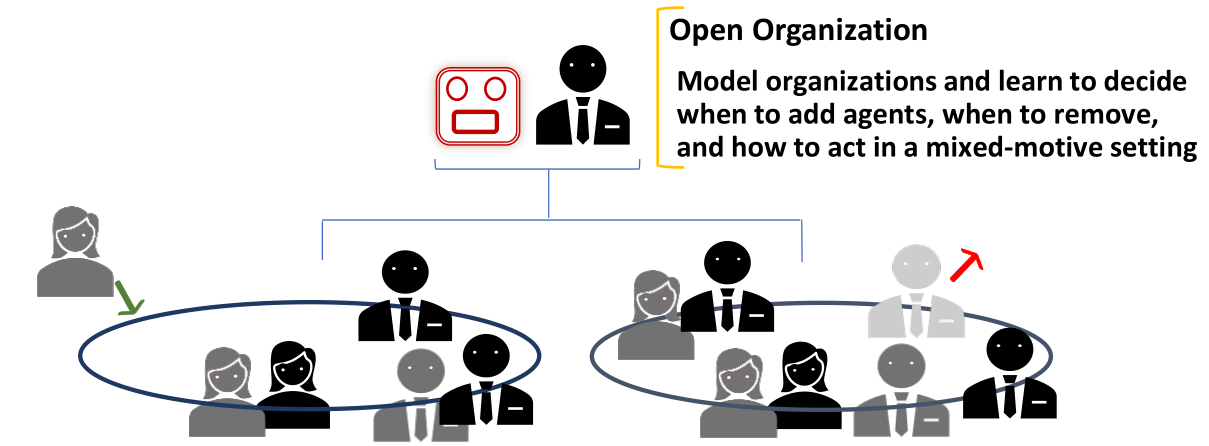}}
    \caption{In the open organization domain, a manager can remove or add an employee to a group while any employee may resign.}
    \label{fig:open_org}
    \vspace{-0.1in}
\end{figure}

\Org{} offers substantially more realistic challenges than previous MARL evaluation domains. However, the original \Org{} assumes a closed system~\citep{ia2c}. In real-life organizations, employees may leave the organization due to various reasons such as retirement or job-hopping. The organization hires employees to fill job vacancies, or lays off employees due to business shrinkage. To simulate this, we introduce the {\em Open \Org{}} domain that generalizes \Org{} to an open multiagent system (see Fig.~\ref{fig:open_org} for a visualization). Open \Org{} contains two types of agents: {\em employee} and {\em manager}. Employees have the option to leave the organization. A manager can hire new employees or fire existing employees based on the current financial health level of the organization. Unlike existing open system domains where external agents enter the system periodically and/or randomly~\citep{openLearning}, in Open \Org{}, the system openness is fully controlled by internal agents. The goal of the internal agents is to optimize their payoff under the influence of system openness.

\subsection{States and Observations}

\begin{figure}[ht!]
    \includegraphics[width=.45\textwidth]{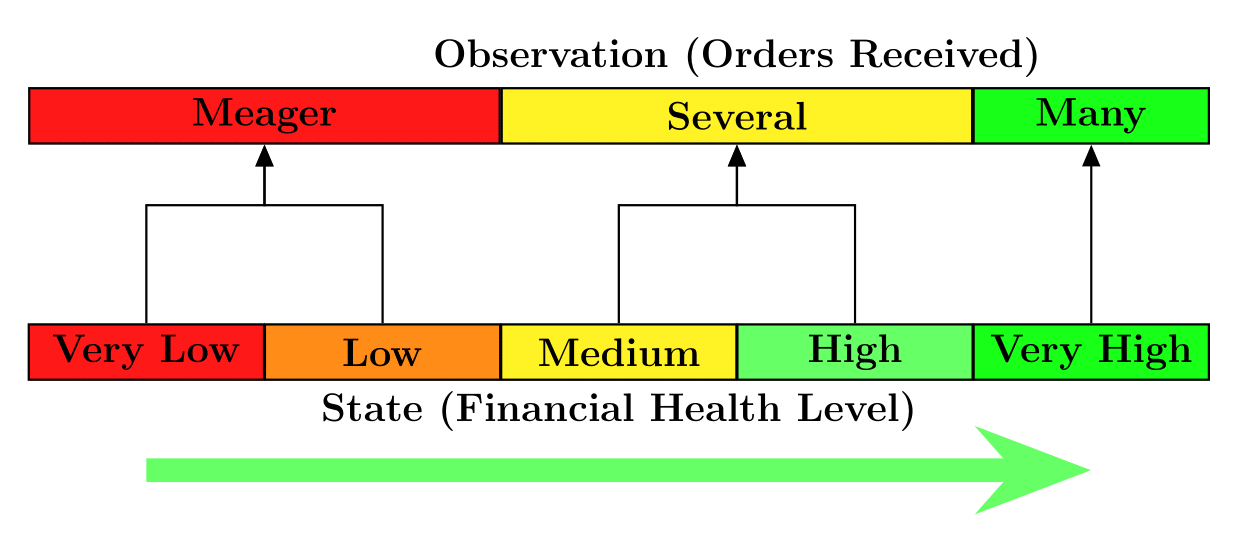}
    \caption{States, observations, and their relationship in Open \Org{}.}
    \label{fig:org}
    \vspace{-0.15in}
\end{figure}

Five states represent the organization's financial health levels as shown in Fig.~\ref{fig:org}: very low ($s_{vl}$), low ($s_l$), medium ($s_m$), high ($s_h$), and very high ($s_{vh}$). States are not observable by agents. Instead, agents receive observations of the number of orders received by the organization that relates to states, and this information is public. Meager ($o_e$) is observed when the organization is in either $s_{vl}$ or $s_l$. Several ($o_s$) is observed when the organization is in either $s_m$ or $s_h$. Many ($o_m$) is observed when the organization is in $s_{vh}$. Agents also receive private observations representing other agent's actions. Private observations reveal other agent's true action with probability $1 - \delta$, where $0\leq \delta<1$.

\subsection{Action and State Transition}

Employees have four possible actions: {\sf self}, {\sf balance}, {\sf group}, and {\sf resign}. {\sf Self} action gives a high individual reward and no group reward; {\sf group} action gives a medium group reward and no individual reward; {\sf balance} action gives a low individual reward and a low group reward. The {\sf resign} action removes the employee from the organization. 
In this paper, we let \Org{} contain a single manager only. The manager has six possible actions: {\sf self}, {\sf balance}, {\sf group}, {\sf fire}, and {\sf hire}. {\sf Self}, {\sf balance}, and {\sf group} are similar to the employee's actions. The {\sf fire} action removes the most recently hired employee, reminiscent of the recent tech sector layoffs as a corrective action to pandemic era overhiring. The {\sf hire} action adds a new employee to the organization.

Joint actions are determined by the number of agents picking {\sf self} compared to those picking {\sf group}, which may transition the state. More agents picking {\sf self} results in lowering the organization's financial health. More agents picking {\sf group} results in improving the organization's financial health. If the same number of agents pick both, the organization's financial health remains unchanged. The employee's action {\sf resign} and manager's action {\sf fire} and {\sf hire} do not impact the financial health of the organization. 
In addition, agents only receive noisy observations of others' actions. {\em To maximize the overall return, an agent needs to trade off individual benefits with group welfare to optimize long-term payoff.}

\subsection{Rewards}

The employee's reward function is the same as in the original \Org{}. The manager's reward function consists of three components: individual, group, and cost:
\begin{equation*}
    R_m^t = \sigma(\sum_i R_i^t) + R_G^t - C^t
\end{equation*}
where $\sigma$ is a Sigmoid function that simulates diminishing returns from the individual rewards of too many employees. $\sum_i R_i$ is the sum of all agents' individual rewards; $R_G$ is the group reward determined by joint action; and $C = c \times \#$ of employees hired in this time step, where $c$ is the cost of hiring one employee. 

\subsection{Agent Openness in Organization}

An open multiagent system is more challenging than a closed one in terms of non-stationarity. This is because any employee agent may leave the system at will or can be removed, and a new employee may enter the system. New agents may have no knowledge about the domain and start with explorational actions, or they may be pre-trained agents with domain knowledge. 

In several of the currently studied open agent domains, the internal agents do not have complete control over the system openness. For example, \cite{openLearning} presents an open system simulation of the highway driving domain. High speed vehicles randomly enter the system, while internal vehicles leave the system if they are too far away from the subjective vehicle. Another example is the firefighting domain introduced in \cite{openPlanning}. Internal firefighters leave the system if they run out of fire suppressants. Only the firefighters who left the system previously can subsequently re-enter. When the internal agents only have minimal control over the system openness, rational decision making entails modeling the openness and predicting the impact caused by it. In contrast, in Open \Org{}, internal agents can  actively manage the system openness in order to reach optimality.  

\section{Latent Interactive A2C}
\label{sec:LIA2C}

In this section, we present \textbf{L}atent \textbf{I}nteractive A2C (LIA2C), which utilizes an encoder-decoder network to model the underlying hidden state and the predicted actions of the agent population as latent variables, for RL in partially observable multiagent environments. 

\subsection{Encoder-Decoder Network}

\begin{figure}[ht!]
    \centering
    \includegraphics[width=.49\textwidth]{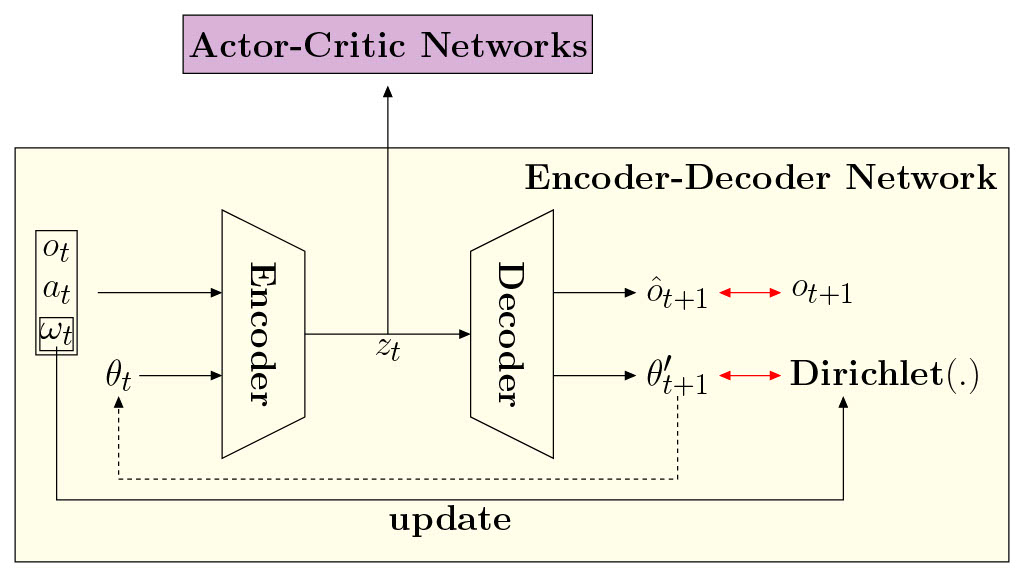}
    \caption{The encoder-decoder network for modeling underlying state and agent population. Here, $\omega_t$ and $o_t$ are the current private and public observations, $a_t$ is the self action, $\theta_t$ is the predicted action distribution by the decoder from time step $t$. $\hat{o}_{t+1}$ is the predicted next public observation, $\theta_{t+1}'$ is the updated action distribution of the agent population. Notice that the original Dirichlet also relies on $\omega_t$ for its update.}
    \label{fig:ed_network}
\end{figure}

We aim to replace the belief filter module of the critic of IA2C$^{++}$ with an encoder-decoder combine. Figure~\ref{fig:ed_network} depicts the neural network architecture of this encoder-decoder. 

At each time step $t$, the inputs to the encoder network are the public observation $o_t$, private observation $\omega_t$, self action $a_t$, and action distribution of the agent population $\theta_t$. The encoder network generates a latent embedding $z_t$ based on the inputs and label. The latent embedding $z_t$ is sent to the actor-critic network for RL, and the decoder network for reconstruction. Given $z_t$, the decoder network generates the public observation $o_{t+1}$ of the next time step through the observation reconstruction head $f_d^o$, and the updated action distribution of the agent population $\theta_{t+1}$ through the model reconstruction head $f_d^\theta$. The loss function of the encoder-decoder networks are defined as,
\begin{align}
    \mathcal{L} = &\frac{1}{H}\sum_{t=1}^H[(f_d^o(z_t)-o_{t+1})^2 - \log Pr(f_d^\theta(z_t)|\omega_t, a_t)\nonumber\\ 
    & + \mathcal{D}_{KL}(Dir(\mathcal{C};f_d^\theta(z_t))||Dir(\alpha+\mathcal{C}';\theta))]\nonumber\\
    = &\frac{1}{H}\sum_{t=1}^H[(f_d^o(z_t)-o_{t+1})^2 - \log Dir(\alpha+{\mathcal C}';f_d^\theta(z_t))\nonumber\\ 
    & + \mathcal{D}_{KL}(Dir(\mathcal{C};f_d^\theta(z_t))||Dir(\alpha+\mathcal{C}';\theta))]
\end{align}
The first term of the loss function is the mean squared error between the reconstructed next observation $\hat{o}_{t+1}$ and the true next observation $o_{t+1}$. The second term is the log posterior of the Dirichlet-multinomial model (see Eq.~\ref{eqn:approx_posterior}) that we want to maximize. ${\mathcal C}'$ is calculated by the rectified method as mentioned in Section~\ref{sec:IA2C$^{++}$}. The last loss term is the KL-Divergence between the reconstructed model and the model obtained by maintaining a Dirichlet distribution using private observation history.
Here $\theta$ in the second term under KL-Divergence is sampled from the distribution $Pr(\boldsymbol{\theta}|\boldsymbol{\alpha+{\mathcal C}'})$ following Eq.~\ref{eq:theta_given_alpha}. The $\mathcal{C}$ in the first term under KL-Divergence is sampled from $Pr(\mathcal{C}|f_d^\theta(z_t))$ following Eq.~\ref{eq:c_given_theta}. Via the last two loss terms, the network can be viewed as learning both to match the external belief update algorithm driven by the Dirichlet-multinomial model (E.~\ref{eq:theta_given_alpha}--\ref{eqn:approx_posterior}) as well as to maximize the resulting posterior. Intuitively, this network {\em learns to update belief} over hidden variables, thus creating a latent representation $z_t$ in the belief space. This latent representation serves as an appropriate augmentation for our decentralized critic.

What would be the consequence if the network did not learn to match the external belief update algorithm, and instead learned its own internal version of such an algorithm? The corresponding loss function would not contain the last loss term, and the posterior maximization (second term) would be the sole driver of this aspect of learning. We conjecture that this version will be slower to converge. Additionally, it may learn similar policies as existing baselines that also do not explicitly model belief update. We include this version as a baseline in our experiments, called LIA2C-w/oKLD, to evaluate these conjectures.

Note that the Dirichlet distributions are only needed for training. During execution, it is not required to maintain a Dirichlet distribution as our actor networks do not use $z_t$ or anything else that this network produces. Hence, both the critic and this encoder-decoder network can be discarded at execution time.


\subsection{Actor-Critic Network for RL}

We augment the input of our critic network to include the latent embedding, $z_t$, from the encoder-decoder network. In particular, the $Q$-function of agent $0$ is now $Q_0(z,o,a_0,\omega_0)$. The advantage function of the critic is defined as,
\begin{align}
    A_0(z,o,a_0,\omega_0) =& ~avg~[r+\gamma~Q_0(z',o'',a'_0,\omega''_0) \nonumber\\
    &- Q_0(z, o', a_0,\omega'_0)]
\end{align}
where $r$, $o'$, $o''$, $\omega'_0$, $\omega''_0$, and $a'_0$ are samples, $z$ and $z'$ are latent embedding of the encoder-decoder network, $avg$ is taken over the sampled trajectories. The agent population's predicted action for the next time step is obtained by sampling from the updated decoder model. Note that the gradients from critic update are not backpropagated to the encoder-decoder network.

It would also be desirable to augment the actor network's inputs to include the latent embedding as it would bring useful information about the hidden state to bear on the actor's  computation. However, we avoid this for two reasons. The first reason is technical difficulty: to run the actor to output $a_t$, we cannot use $z_t$ as one of its inputs because the encoder itself requires $a_t$ as one of its inputs. Using $z_{t-1}$ as an actor input instead would  introduce a one-step lag. The second reason is historical precedent. As~\cite{slac} argue in their paper, ``... the policy is not conditioned on the latent state, as this can lead to  over-optimistic behavior since the algorithm would learn Q-values for policies that have perfect access to the latent state. Instead, the learned policy in our algorithm is conditioned directly on the past observations and actions. This has the additional benefit that the learned policy can be executed at run time without requiring inference of the latent state.'' This observation applies to LIA2C as well. Consequently, we leave the actor network to recommend an action based solely on the public observation $o$. Note that the latent state still has an impact on the actor's updates via the advantage term, as its gradient is
\begin{equation}
    avg~[\nabla_\theta \log {\pi_\theta}(a_0|o)~A_0(z,o,a_0,\omega_0)].
\end{equation}

\begin{figure}[t!]
    \centering
    \includegraphics[width=.49\textwidth]{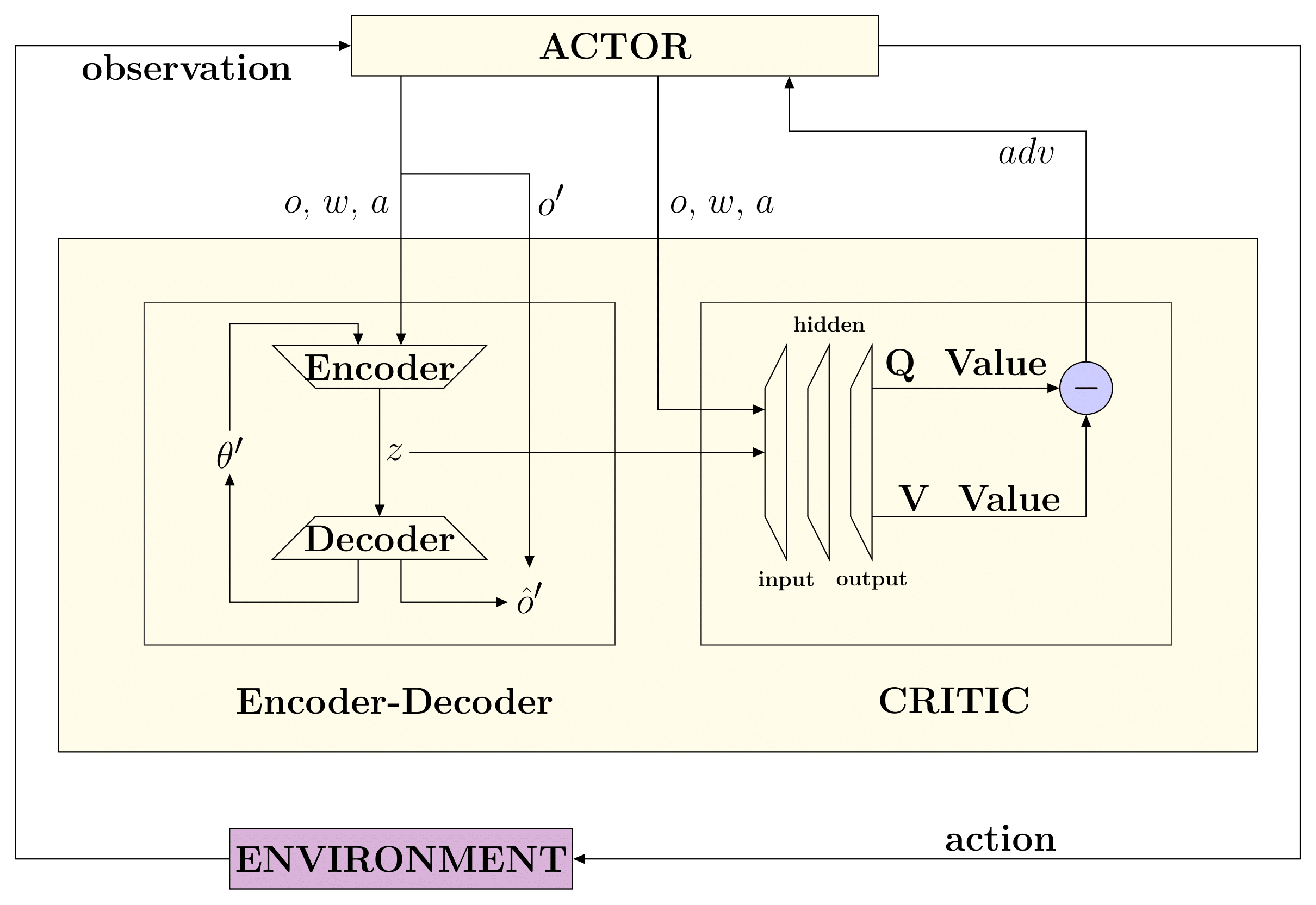}
    \caption{Overall architecture of LIA2C. The encoder-decoder network updates the model of the agent population. The critic's advantage function now depends on the latent embedding, self action, and observations instead of the configuration.}
    \label{fig:network}
    \vspace{-0.2in}
\end{figure}

Figure~\ref{fig:network} demonstrates the overall network architecture of LIA2C. The actor receives observation from environment and sends self actions to the encoder-decoder network. The encoder-decoder network updates the approximated model $\theta'$ based on the actor's action and observations from environment. The latent embedding of the encoder-decoder network is sent to the critic for advantage computation. The critic network updates its parameters based on the latent embedding and environment reward, and sends the advantage value to the actor for its gradient update.

\section{Experiments}
\label{sec:experiment}

We implemented LIA2C in Python~\footnote{Our code will be released upon acceptance.} and evaluate its learning performance on multiple instances of two domains in reference to significant baselines.

Our first experimental domain is the \Org{}; we use the original (closed) \Org{}~\citep{mia2c} and our new Open \Org{} (described in Section~\ref{sec:org}). Our \Org{} is fully connected, where all agents share one neighborhood. For the Open \Org{} domain, the cost $c$ for hiring one employee is set to $1$, and the individual reward for {\sf resign}, {\sf hire}, and {\sf fire} are all set to $0$. The environment initiates with one manager and one employee. 

\begin{figure}[ht!]
\center
    \begin{subfigure}{.2\textwidth}
    \includegraphics[width=1\linewidth]{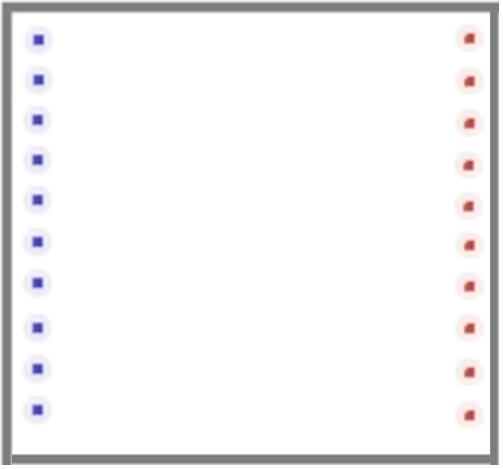}
    \caption{initial position}
    \label{fig:MA1}
    \end{subfigure}
    \begin{subfigure}{.2\textwidth}
    \includegraphics[width=1\linewidth]{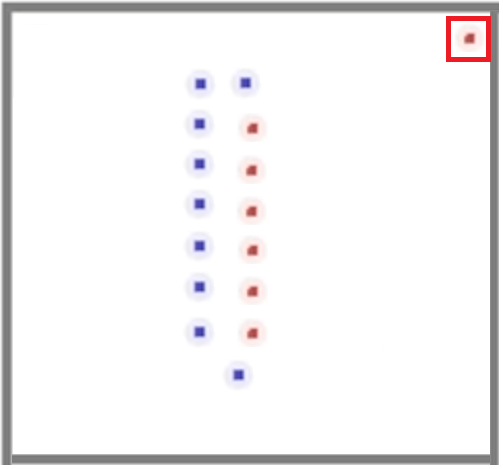}
    \caption{re-spawned red agent}
    \label{fig:MA2}
    \end{subfigure}
    \caption{($a$) Agents from blue team are deployed on the left side, while agents from red team are deployed on the right side. ($b$) A red agent re-spawned after {\sf retreat}.}
    \label{fig:battle}
\end{figure}

Our second experimental domain is the Battlefield~\citep{ia2c} in \Mag{} environment~\citep{magent} with 10 agents from each team as shown in Fig.~\ref{fig:battle}. We introduce system openness to the Battlefield by adding a new action {\sf retreat}. All blue agents are spawned at the left side of the field and all red agents are spawned at the right side of the field. If the agent was not eliminated at the current time step, the {\sf retreat} action removes the agent from the current location, and re-spawns the agent at the left side or right side depending on its color. 

\paragraph{Baselines} Baseline methods are LIA2C without the KL-Divergence term in loss function labeled as LIA2C-w/oKLD, IA2C$^{++}$DM~\citep{mia2c}, and a recent method for modeling other agents that also uses an encoder-decoder labeled as LIAM~\citep{liam}. Implementation details and hyperparameters are specified in Appendix A.

\subsection{Model Performance}
\begin{figure}[ht!]
    \centering
    \begin{subfigure}{.25\textwidth}
    \includegraphics[width=1\linewidth]{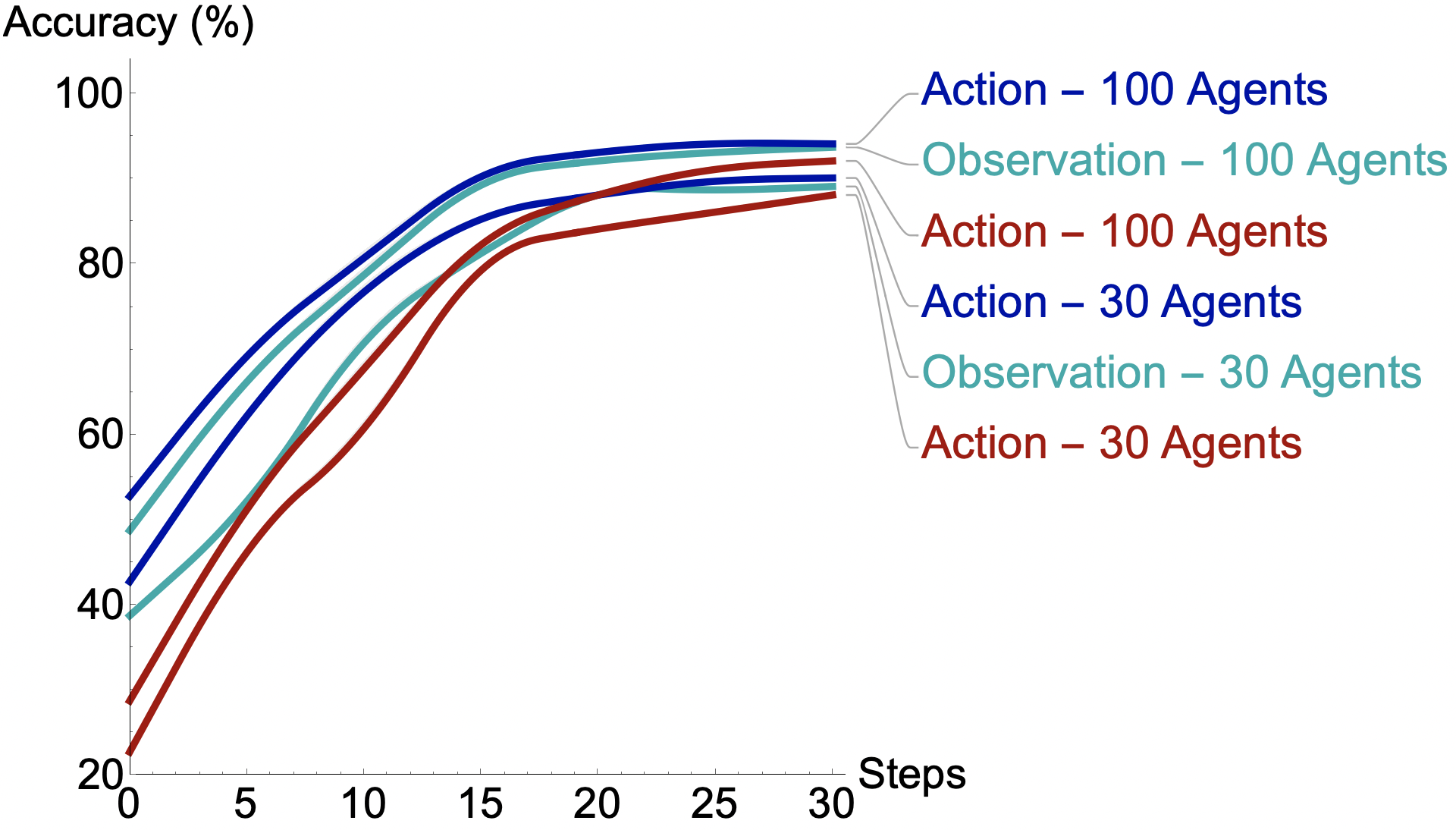}
    \caption{Prediction accuracy}
    \label{fig:accuracy}
    \end{subfigure}
    \begin{subfigure}{.22\textwidth}
    \includegraphics[width=1\linewidth]{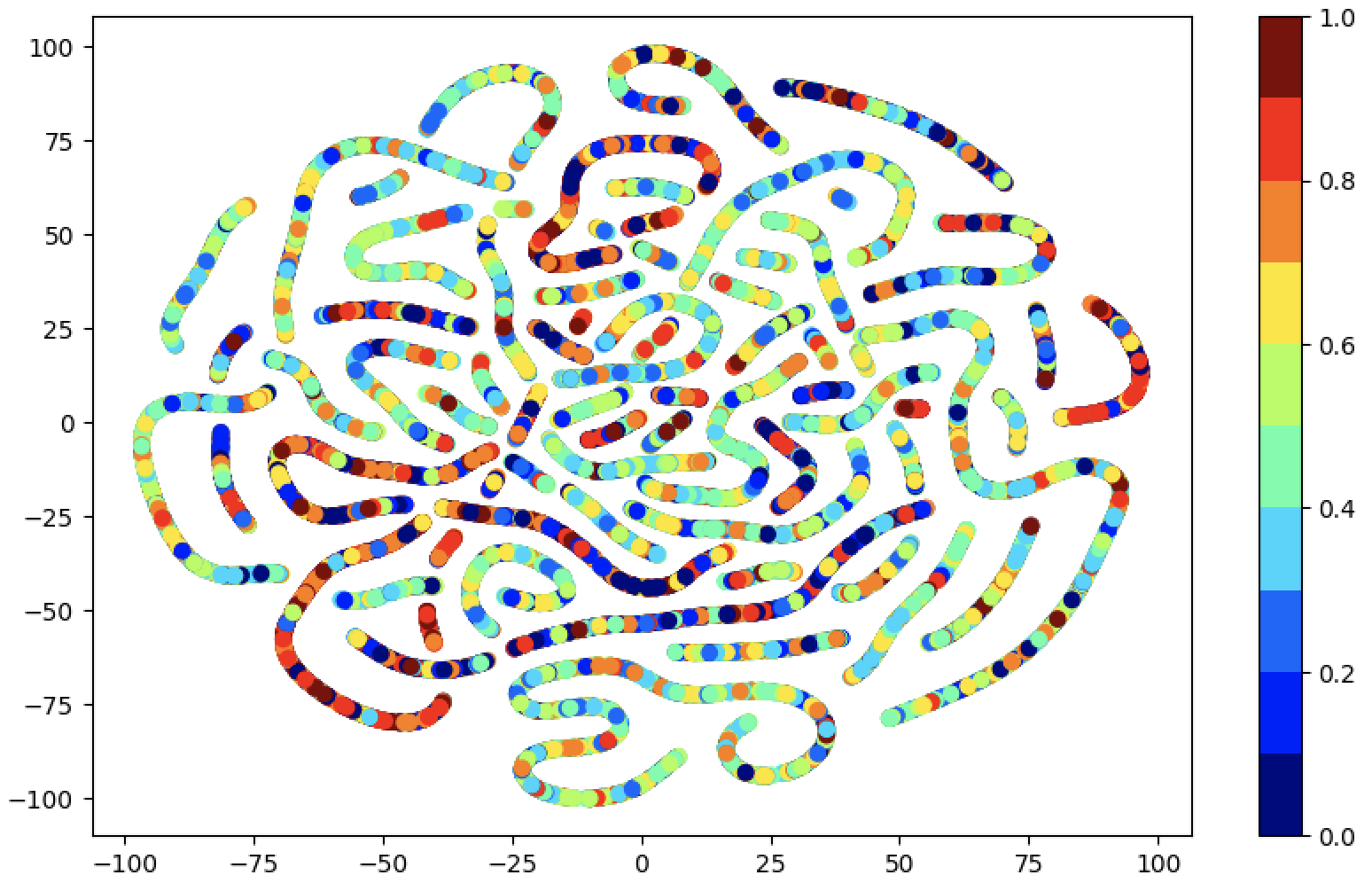}
    \caption{t-SNE plot for latent embeddings}
    \label{fig:tsne}
    \end{subfigure}
    \caption{(a) Action and observation prediction accuracy with respect to time steps. Blue and cyan lines are the prediction accuracy of LIA2C, and red lines are the prediction accuracy of IA2C$^{++}$DM. (b) The t-SNE plot for the encoder-decoder latent embeddings. Different colors represent different distributions of $\theta$ and different points represent different time steps.}
    \label{fig:model_performance}
\end{figure}

\paragraph{Prediction accuracy}
We evaluate the performance of the encoder-decoder model by its prediction accuracy. Figure~\ref{fig:accuracy} shows the prediction accuracy of actions (blue) and observations (cyan) for the encoder-decoder model in LIA2C, and the prediction accuracy of actions (red) for the Dirichlet model in IA2C$^{++}$DM.  
\paragraph{Visualization}
We visualize the latent embeddings learned by the encoder-decoder model using t-SNE~\citep{tsne}. Figure~\ref{fig:tsne} shows the two-dimensional projection of the latent embeddings. We observe that points with similar colors are connected, indicating a smooth transition of the belief over others' action distribution.   

\subsection{Comparative Performance} 

\begin{figure}[ht]
\center
    \begin{subfigure}{.20\textwidth}
    \includegraphics[width=1\linewidth]{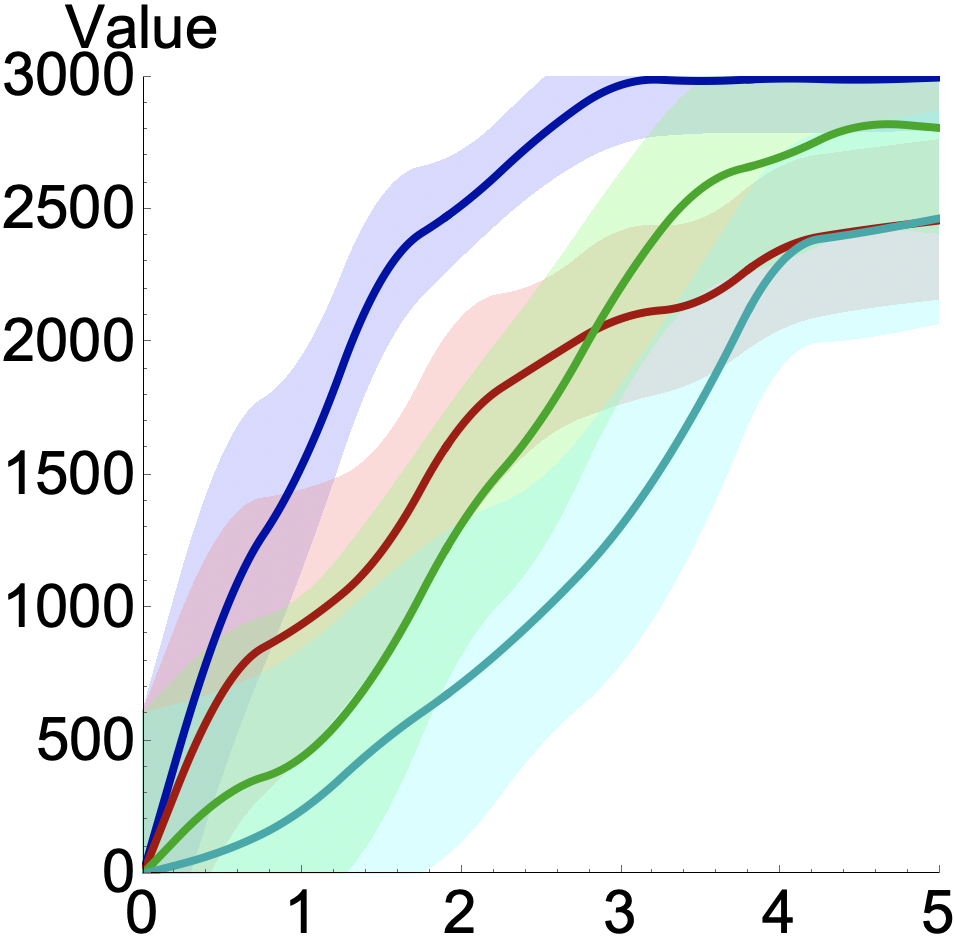}
    \caption{30 agents \Org{}}
    \label{fig:org30}
    \end{subfigure}
    \begin{subfigure}{.26\textwidth}
    \includegraphics[width=1\linewidth]{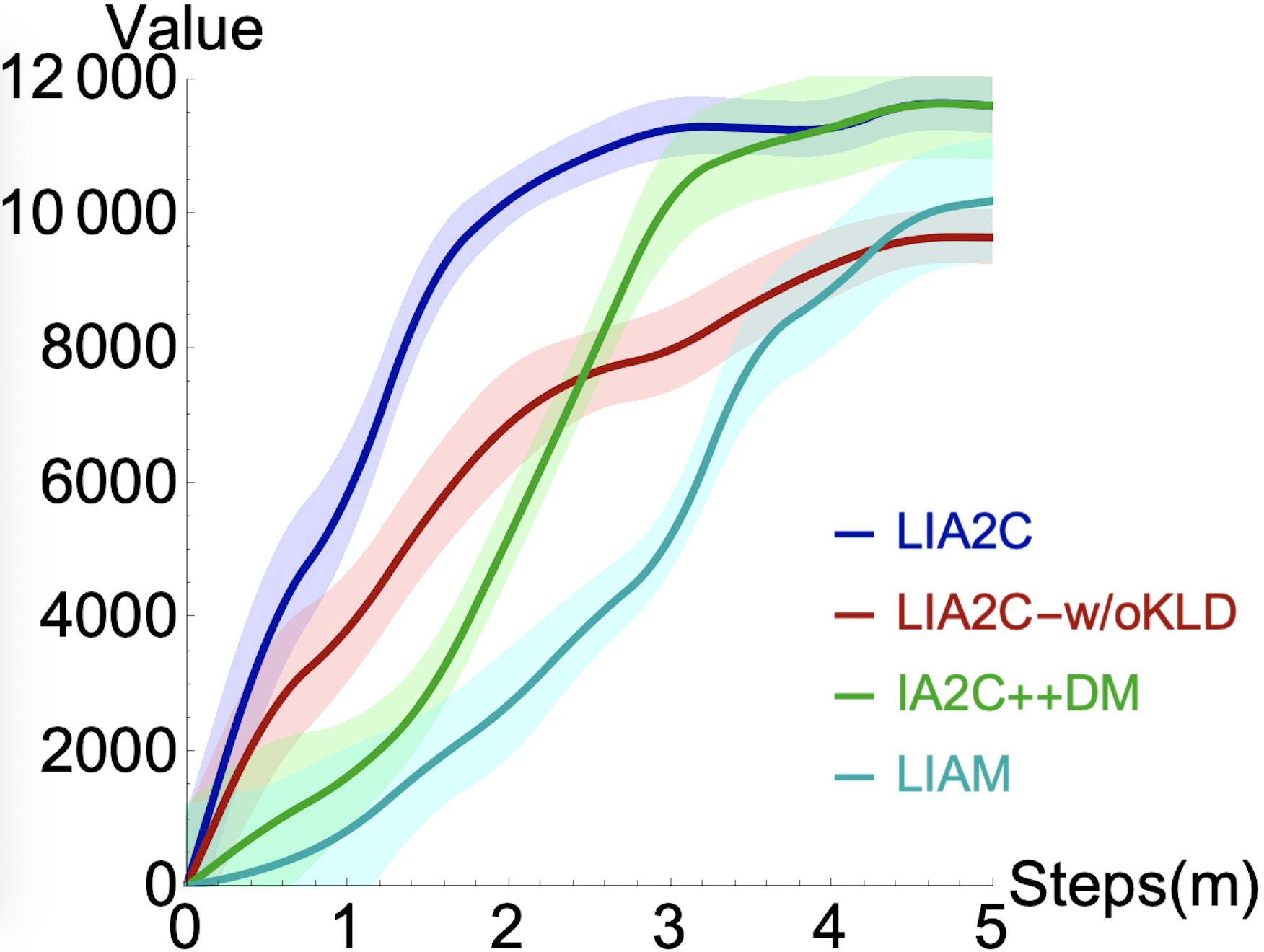}
    \caption{100 agents \Org{}}
    \label{fig:org100}
    \end{subfigure}
    \caption{(a) The variance of LIA2C is lower than IA2C$^{++}$DM and LIAM. (b)  LIA2C is more sample efficient as it requires much fewer samples to converge than baseline methods.}
    \label{fig:org_results}
\end{figure}

\paragraph{\Org{} and Open \Org{}} Figure~\ref{fig:org_results} shows the experimental result from the original  \Org{} with 30 and 100 agents. In the original \Org{}, LIA2C converges to the optimal policy in roughly 2M steps. IA2C$^{++}$DM converges to the optimal policy after 3.5M steps. LIA2C-w/oKLD does not converge to the optimal policy within 5M steps. Due to the noisy private observations, LIAM only converges to sub-optimal policies. In addition, the averaged variance of LIA2C is 32$\%$ lower than IA2C$^{++}$DM, and 44$\%$ lower than LIAM. 

\begin{figure*}[ht!]
    \centering
    \begin{subfigure}{.28\textwidth}
    \includegraphics[width=1\linewidth]{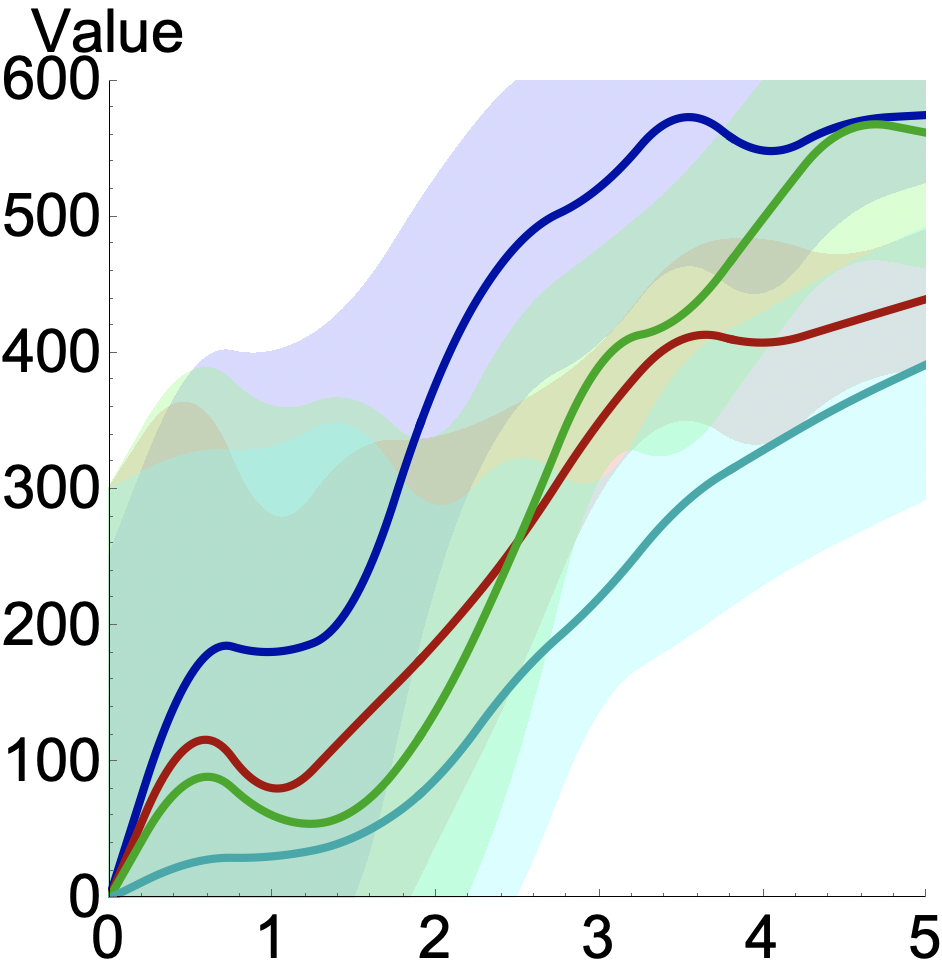}
    \caption{Open \Org{}}
    \label{fig:openorg_results}
    \end{subfigure}
    \begin{subfigure}{.30\textwidth}
    \includegraphics[width=1\linewidth]{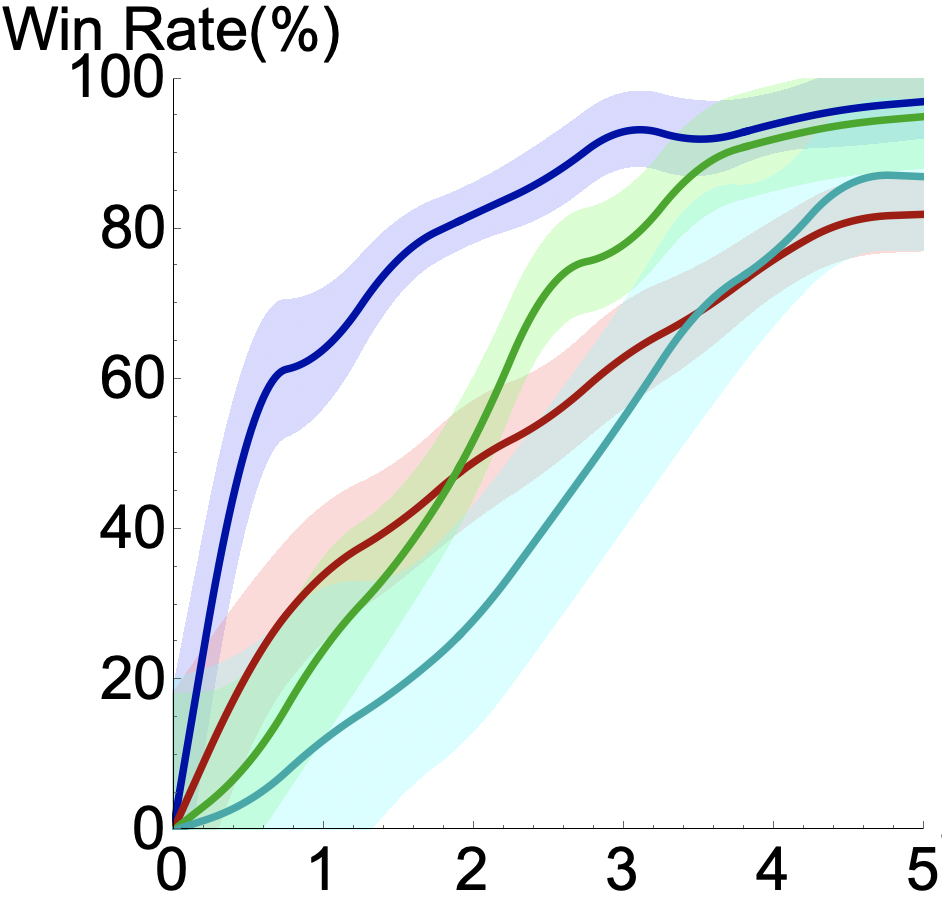}
    \caption{\Mag{} battle field}
    \label{fig:magent_results}
    \end{subfigure}
    \begin{subfigure}{.38\textwidth}
    \includegraphics[width=1\linewidth]{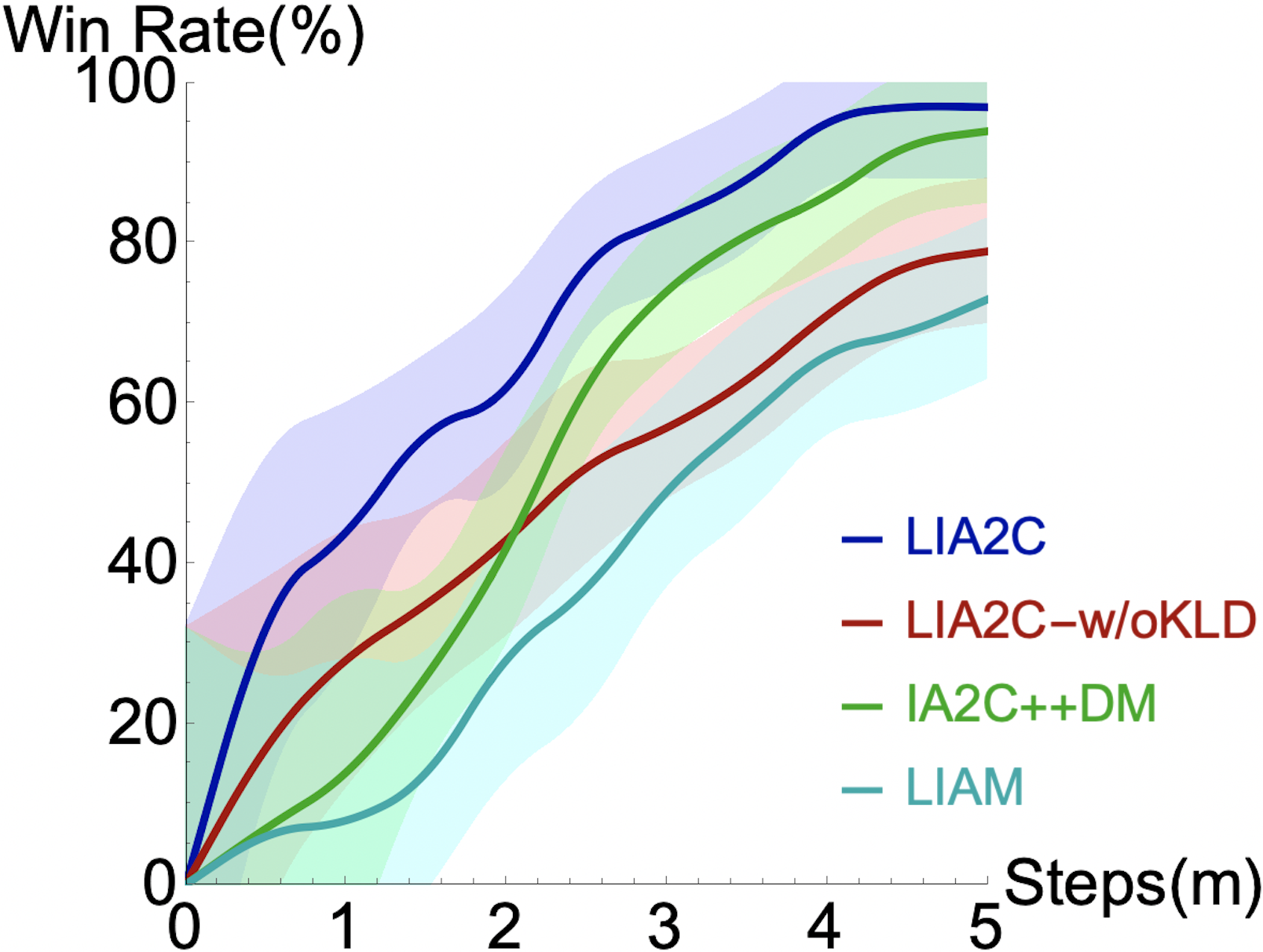}
    \caption{Open \Mag{} battlefield}
    \label{fig:openmagent_results}
    \end{subfigure}
    \caption{(a) Open \Org{} starts with one manager and one employee. LIA2C and IA2C$^{++}$DM policies hire a total of 8 employees, which lead to optimal payoff for the manager. (b) After 2M steps, LIA2C learned a policy that has around 80$\%$ win rate, while baseline methods policies’ win rates are all less than 50$\%$. (c) All methods took more steps to learn high-quality policies due to the additional uncertainty introduced by system openness.}
    \label{fig:results}
\end{figure*}

Experiment in the open \Org{} starts with 1 manager and 1 employee. The manager's individual reward component is discounted by $0.9^{E-1}$, where $E$ is the number of employees hired. The manager's cost $C$ for hiring an employee is 1 per time step. Figure~\ref{fig:openorg_results} shows policy values of the different methods in the open \Org{}.

We observe higher variances from all methods in open \Org{}. The system openness has greatly exacerbated the non-stationarity of the environment. LIA2C converges after 3.5M steps, while IA2C$^{++}$DM converges after 4.5M steps. Both LIA2C and IA2C$^{++}$DM learn to hire 8 employees, which is the optimal number of employees that should be hired according to the reward function. An employee's policy remains the same as the original \Org{}. LIA2C-w/oKLD and LIAM did not converge to high-quality policies. 

\paragraph{\Mag{} Battlefield and Open \Mag{} Battlefield} In the \Mag{} Battlefield domain, all methods are tested against agents pre-trained by independent A2C. Figure~\ref{fig:magent_results} shows averaged win rate of corresponding methods. 

LIA2C learns high-quality policies after 2M steps. IA2C$^{++}$DM learns similar policies after 4M steps. LIAM converges to policies that are close to LIA2C and IA2C$^{++}$DM, but the variances are higher. On average, the variance of LIA2C is 29$\%$ lower than IA2C$^{++}$DM, and 41$\%$ lower than LIAM. 

In the open \Mag{} Battlefield domain, all methods are tested against agents pre-trained by independent A2C. Figure~\ref{fig:openmagent_results} shows averaged win rate of corresponding methods. 

Due to the system openness, it takes LIA2C nearly 4M steps to converge to high-quality policies. IA2C$^{++}$DM learns policies that are very close to LIA2C. We notice a drop in the performance of LIAM in the open \Mag{} Battlefield. LIAM appears to struggle with learning accurate latent representations in open systems in the presence of noise in private observations. 

\section{Related Work}\label{sec:relatedwork}
\paragraph{Multiagent reinforcement learning}
CTDE has been a common paradigm in recent MARL research. Prior works~\citep{maddpg,coma,mappo} adopt a centralized critic and decentralized actor architecture. Another thread of investigation within CTDE has been value-decomposition to alleviate joint action based value representation. \citet{vdn,qmix} decompose joint Q-functions as a function of individual agents local Q-functions. The use of unsupervised learning techniques to learn low-dimensional representations of the environment has led to promising improvements in policy quality and sample efficiency. Recent works~\citep{side, liam} focus on learning variable models which are the joint of state representations and environment dynamics. In contrast to these works, LIA2C focuses on learning representations of the interactive state as well as latent state dynamics.

\paragraph{Planning and learning in multiagent open systems}
In most real-world problems, the number of agents in the environment is usually not fixed. Agents can leave or enter the environment due to various reasons. The system openness together with multiagent setting have greatly increased the non-stationarity of the environment. \citet{openPlanning,newopen} model multiagent open systems from a decision-theoretic perspective. The uncertainty in multiagent open systems is addressed by modeling neighbors and predicting their future presence and behavior. Transfer learning approaches can also be used for RL in multiagent open systems. \citet{openLearning} proposed a CTDE method by learning a single-agent policy and then transferring knowledge from a single-agent setting to a multiagent setting. The policy is shared by direct parameter sharing and hence only applicable in homogeneous populations. A recent method called graph-based policy learning (GPL)~\citep{openlearn} builds on graph neural networks (GNN) to learn agent models and joint-action value models in fully observable multiagent open systems. GPL learns joint action-value function that models the effects of other agents' actions towards the subjective agent's returns, along with a GNN-based model to predict the actions of other teammates. 

\vspace{-0.1in}
\section{Concluding Remarks}
\label{sec:conclusion}
\vspace{-0.05in}
We have presented a new method for reinforcement learning in many-agent systems in a DTDE setting, called LIA2C. This method extends a current trend of latent space modeling for MARL, but contrasts with the existing literature in an important way. Instead of learning a latent representation of merely the hidden variables, LIA2C's latent space encompasses both the variables and a belief update algorithm over such variables. The resulting enhancement to the model's predictive power translates to higher sample efficiency as well as relatively lower variance compared to recent baselines in two many-agent domains---\Org{} and \Mag{}, along with their open variants that we have also introduced in this paper. Additionally, from the observation that the quality of LIAM's learned policies are not too different from LIA2C-w/oKLD's, we conclude that explicitly modeling belief update---something that LIAM also does not---is a critical factor behind LIA2C's success. 

This study opens up exciting directions for future investigation. The high variance observed in our experiments in open domains indicates a need for renewed attention to non-stationarity, long masked by focus on CTDE. Indeed, volatility introduced by openness may very well be amenable to latent space modeling. While we have adopted existing MARL modeling concepts such as Dirichlet multinomial model and mean field action for this paper, other concepts that leverage the rich literature on statistical mechanics (e.g., see~\citep{Hu19:Modeling}) may be especially suited for open many-agent systems, and may benefit future investigation.

\bibliography{hdbUAI23}
\end{document}